\PassOptionsToPackage{table}{xcolor}
\documentclass{article} %
\usepackage{iclr2027_conference,times}

\usepackage{amsmath,amsfonts,bm}

\def\eqref#1{equation~\ref{#1}}

\def\1{\bm{1}}

\DeclareMathAlphabet{\mathsfit}{\encodingdefault}{\sfdefault}{m}{sl}
\SetMathAlphabet{\mathsfit}{bold}{\encodingdefault}{\sfdefault}{bx}{n}

\usepackage{hyperref}
\usepackage{url}
\usepackage{fontawesome5}
\usepackage{xcolor}
\usepackage{array}
\usepackage{xfp}
\usepackage{amssymb}
\usepackage{graphicx}
\usepackage{booktabs}
\usepackage{multirow}
\usepackage{subcaption}
\usepackage{wrapfig}
\usepackage[most]{tcolorbox}
\usepackage{listings}
\usepackage[normalem]{ulem}
\usepackage{xspace}
\usepackage{xurl}

\definecolor{dropmarkcolor}{HTML}{D62728}
\newcommand{\dropmark}{\textcolor{dropmarkcolor}{$\blacktriangledown$}}

\newcommand{\native}{native\xspace}
\newcommand{\shared}{shared\xspace}

\newcommand{\sysname}{\textsc{BlindToChange}\xspace}

\title{BlindToChange: AI Agents Can Find Changes When Asked But Do They Look When Needed?}

\author{Anirudh Phukan \\
Indian Institute of Science\\
\texttt{anirudhp@iisc.ac.in} \\
\And
Tanya Goyal \\
Cornell University\\
\texttt{tanyagoyal@cornell.edu} \\
\And
Danish Pruthi \\
Indian Institute of Science\\
\texttt{danishp@iisc.ac.in}
}

\iclrfinalcopy %
\begin{document}

\maketitle
\lhead{Preprint}

\begin{abstract}

Given that AI agents can search the web broadly and deeply in real time, it is tempting to conclude that their outputs account for recent world events. We show that this conclusion does not hold in practice. While agents \textit{can} search and easily find recent changes, the correctness of their responses depends on \textit{whether} they search and \textit{what} they search for. Popular agentic search benchmarks like BrowseComp cannot expose failures wherein agents do not recognize their own assumptions that should be verified, since these benchmarks explicitly state what needs to be found. In contrast, many user requests state only the final information need (e.g., an itinerary for a trip), leaving the agent to discover relevant changes (e.g., the temporary closure of a museum) and account for their consequences.
To highlight such gaps, we introduce \sysname, a benchmark that measures whether agents proactively discover recent events that affect their responses. Our benchmark comprises requests for recommendations where an obvious candidate is no longer applicable due to a recent real-world event. These events are drawn automatically from Wikidata, allowing \sysname to be periodically refreshed.
Strikingly, we 
find a big gap---as high as 27.5 points, even at high reasoning effort---in systems' 
performance, 
when the recent event is not explicitly mentioned compared to when we ask about it directly.
Analyzing search trajectories, 
we learn that more than half of the failures can be attributed to 
the fact that agents 
do not directly search to verify whether a relevant fact has changed.
Our benchmark 
highlights, and tracks, 
a new axis 
of progress, 
wherein systems
account for  
recent  
information 
that is not 
explicitly mentioned in the user request.

\end{abstract}

\begin{center}
\vspace{-2mm}
{\small
\faGithub\ \href{https://github.com/Anirudh-Phukan/blind-to-change}{\texttt{Anirudh-Phukan/blind-to-change}}\quad
\raisebox{-0.15em}{\includegraphics[height=1em]{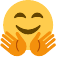}}\ \href{https://huggingface.co/datasets/Anirudh-Phukan/blind-to-change}{\texttt{Anirudh-Phukan/blind-to-change}}}
\end{center}

\begin{figure}[htbp]
\centering
\includegraphics[width=0.93\linewidth]{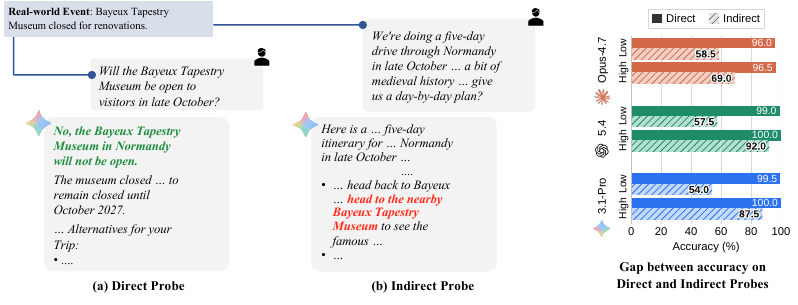}
\caption{Two probes for the same underlying update that the Bayeux Tapestry Museum is closed until October 2027, posed to \texttt{gemini-3.1-pro} with search enabled. In (a) the request names the museum and asks about its status. The model finds out that the museum is closed and adjusts its recommendation. In (b) the request asks for an itinerary and never mentions the museum. The model recommends it as an afternoon stop and does not check whether it is open. The direct probe shows that the model can find the fact. The indirect probe evaluates whether it discovers the change on its own. The plot on the right shows the gap between direct and indirect probe accuracy.}
\label{fig:probe_example}
\end{figure}

\section{Introduction}
\vspace{-4mm}

AI agents, specifically LLMs equipped with search and browsing tools, are increasingly used for information seeking \citep{chatterji2025people, xi2025survey}. Access to real-time news articles and web pages through these tools invites the conclusion that agents' responses always reflect the current state of the world. We argue that this is not the case, making it all the more concerning that users often rely on these responses without verifying against the cited sources \citep{pew2025aisummaries}.

Consider a user who asks an agent to plan their trip to Normandy (see Figure~\ref{fig:probe_example}b). While the Bayeux Tapestry Museum would normally be an obvious recommendation, it is currently closed for renovations. The agent (Gemini-3.1-pro in the example in the Figure) fails to account for this real-world update and erroneously includes the museum in the user's itinerary. Crucially, such updates are easy for agents to find on the web. When directly asked to verify, the same agent readily informs the user that the museum is closed and suggests alternatives (Figure~\ref{fig:probe_example}a). This inconsistency shifts the burden of verification to end users. Ideally, reliable agents must output well-researched and up-to-date responses by default. Agents must be able to identify unstated constraints (e.g. a museum's current status) and account for them, regardless of whether the user's query explicitly refers to them or asks to verify each fact. Without this, agents risk misleading users who trust their responses at face value.

Recent benchmarks, such as BrowseComp \citep{wei2025browsecomp, chen-etal-2026-browsecomp} or DeepSearchQA \citep{gupta2026deepsearchqa}, also investigate agentic search capabilities but on an orthogonal axis. They require complex long-horizon search but all the constraints that need to be satisfied are clearly specified. Similarly, benchmarks for temporal freshness \citep{vu-etal-2024-freshllms, meem-etal-2024-pat,kim-etal-2024-carpe} directly state what to search for or even include temporal markers (e.g., current) that suggest that facts may have changed. In our setting, on the other hand, relevant changes that impact the final answer are easy to find but the need to find them is not apparent from the query itself. 

We introduce \textsc{BlindToChange}, a framework to study whether search agents proactively discover recent events that affect their responses and account for their consequences. We measure this as the gap between agents' performance on two complementary probes of the same recent event: a direct probe that explicitly states the event in the query and an indirect probe that does not but requires reasoning about its consequences to answer correctly. This framing allows us to cleanly measure agents' ability to discover relevant events and reason about their impact, without confounding our analysis with the difficulty of finding the event itself.

\textsc{BlindToChange} is constructed using an automated pipeline that extracts real-world events from Wikidata and leverages frontier LLMs to construct direct and indirect probes. Given a recent event, direct probes are easy to construct and responses to them are easy to evaluate. We phrase indirect probes as recommendation requests, carefully constructed such that the target event disqualifies an obvious candidate (e.g., the museum in our example). Then, we evaluate the agents' responses for the \textit{absence} of the disqualified candidate. This allows us to objectively score long-form responses to a subjective query. We borrow this idea of constructing indirect probes as recommendation requests from  \cite{li-goyal-2025-memorization}. Their own framework is limited in scale as it was manually curated. Moreover, it supports evaluation only on a small retrieval corpus of synthetically generated events and data, which cannot match the complexity or scale of real-world news and internet data. Therefore, it is inadequate for evaluating frontier agentic search systems. 

We make two key design decisions that ensure the validity of our evaluation framework. First, we restrict events to those that occurred after the reported knowledge cutoff of the agent. Second, we only retain pairs with indirect probes for which the agent included the disqualified candidate in their responses before the event. We approximate behavior before the event by sampling multiple outputs with search disabled. Together, these ensure that search is necessary to provide accurate answers for both direct and indirect probes, and that the \textit{absence} of the disqualified candidate in a response is due to agentic search and not chance.

Our current dataset comprises direct and indirect probes based on two event types: demises and ends of official tenures of public figures. These were chosen as they have sufficient coverage in Wikidata in the months since the knowledge cutoff of the frontier models, although our framework extends to other event types. We evaluate 8 frontier models in two search harnesses, each model's own native search and a common scaffold with a shared web-search tool, at low and high reasoning effort. The native harness reflects how users actually encounter these systems, while the shared harness fixes the search backend across models and exposes the full search trajectory for analysis.

We find a large gap between what agents can find when asked explicitly and what they discover on their own.  Agents almost always find the change when asked directly at high effort, reaching at least 96.5\% direct accuracy on demises in nearly all configurations. Yet when queried with indirect probes for demises, the accuracy falls by a median 40.5 points at low effort and 27.5 points at high effort under the \native harness; agents routinely recommend people who have passed away. %
The drop persists on tenures but is smaller, with median drops of 18.7 and 15.0 points for low and high reasoning effort, likely because officeholders change routinely and models are more inclined to check whether someone still holds a position. %
Strikingly, our error analysis under the \shared harness finds that many failures stem from agents never verifying their own assumptions rather than from failing to retrieve the change. In fact, this accounts for over 75\% of demise failures in 13 of the 15 configurations. Even on tenures, where models more often check whether the person still holds the position, not searching explicitly accounts for a median of 50\% of failures across configurations. 

One key finding is that strong performance on existing agentic benchmarks does not carry over to \sysname. Notably, Claude-Opus-4.7 has the lowest indirect accuracy on demises under the \shared harness at both effort levels. The issue is not search quality. When it searches as often as other models do, it is nearly as accurate. It simply searches less, issuing no search on 23\% of demise probes at high effort (this number for other models is less than 8\%). On tenures, where models search more readily, it is within 3 points of the most accurate model at both effort levels. GPT-5.4 shows the same pattern at low effort, trailing the open-weights models on demises by 34.0 to 52.5 points on a common set of probes as it issued no search for 58.5\% of them. At high effort, it searches on 97.5\% of demise probes and performs better than all other models.

Taken together, our work shows that finding information and recognizing when to look for it are distinct capabilities. Benchmarks that state what needs to be found therefore overestimate how well agents keep up with a changing world. Closing this gap requires systems that identify which of their own assumptions may be out of date and verify them without being asked. \vspace{-2mm}

\section{Related Work}
\vspace{-4mm}
This section summarizes notable related work. Extended discussion can be found in Appendix \ref{app:related-work}.

\textbf{Agentic Search Benchmarks.} BrowseComp is a popular benchmark for agentic web search, with questions whose evidence is deliberately hard to locate, so answering requires searching both deeply and broadly \citep{wei2025browsecomp}. Other benchmarks directly score how the agent searches. DeepSearchQA \citep{gupta2026deepsearchqa} measures whether agents find a complete set of answers and stops at the right time, InfoDeepSeek \citep{xi2025infodeepseek} scores the quality of the information it gathers, and Mind2Web 2 \citep{gou2025mind2web} checks whether its answers are attributed to sources. LiveBrowseComp and LiveNewsBench separate search from recall by building questions from low-salience or recent facts that a model cannot have memorized \citep{fan2026livebrowsecomp,zhang2026livenewsbench}.

\textbf{Temporal Freshness Benchmarks.} These tests models on questions whose answers change over time. RealTimeQA releases a new question set every week, drawn from news quizzes \citep{kasai2023realtime}. FreshQA has annotators re-verify answers every week to a fixed question set \citep{vu-etal-2024-freshllms}. To reduce manual effort, later benchmarks construct datapoints automatically from sources that record their own changes: PAT-Questions from time-dependent Wikidata relations \citep{meem-etal-2024-pat}, and EvolvingQA, GrowOVER, EvoWiki and DynaQuest from differences between successive Wikipedia snapshots \citep{kim-etal-2024-carpe,ko-etal-2024-growover,tang-etal-2025-evowiki,lin-etal-2025-dynaquest}.

However, both kinds of benchmarks explicitly state what information needs to be found, while we study whether agents discover information that the request never explicitly mentions.

\textbf{Implicit Constraint Benchmarks.} Task success often depends on factors the user never explicitly states but expects an assistant to recognize \citep{kaur-etal-2026-proper}. Several benchmarks make such factors discoverable through tool use. Implicit Intelligence hides constraints in the environment \citep{sirdeshmukh2026implicit}, DeepPlanning checks travel and shopping plans against implicit constraints such as limited seat availability \citep{zhang-etal-2026-deepplanning}, and MapSatisfyBench recovers the decision factors users left unspoken from behavior logs \citep{bai2026mapsatisfybench}. All three find that agents rarely probe for the missing information, even when a tool call would reveal it. We show that implicit constraints also arise when seeking information about a changing world, and find that agents similarly fail to search for and discover them using web search. %

\newcommand{\highlight}[1]{{%
  \setlength{\fboxsep}{1pt}%
  \colorbox{red!15}{#1}%
}}

\section{\texorpdfstring{\sysname}{BlindToChange}}
\vspace{-3mm}
We have the following desiderata for \sysname: (i)~It must measure whether agents proactively discover and account for updates that affect the correctness of their answers; (ii)~its evaluation must be objective and reliable; and (iii)~it must be automatically constructible, and hence easy to scale and refresh. We first describe the conceptual ideas and design choices that help realize these desiderata in \S\ref{sec:indirect-probing}. We then describe our automated pipeline for constructing the benchmark in \S\ref{sec:construction}.

\subsection{Conceptual Framework}
\label{sec:indirect-probing} 
\vspace{-2mm}
\paragraph{Setup.}
Consider an event that changes a fact about an entity $x$ from $f_\text{old}$ to $f_\text{new}$. In Figure~\ref{fig:probe_example}, $x$ is the \textit{Bayeux Tapestry Museum} and the event is its closure. The fact $f_\text{old}$ is that the museum is open to visitors, and $f_\text{new}$ is that it is closed until October 2027. While the event likely changes other facts too, we focus only on the update $u = (f_\text{old}, f_\text{new})$. 

We can represent the knowledge state of the world before and after the event as $K_\text{old}$ and $K_\text{new}$ respectively. Both knowledge states contain a combination of common facts that are unchanged by the event, as well as facts that are outdated (in $K_\text{old}$) or updated (in $K_\text{new}$). 

For a given question $q$, let $\mathcal{A}(q;K)$ denote the set of valid answers under knowledge state $K$. We use a set because many questions, such as requests for recommendations, admit several valid answers. We then define the \textit{probes for an update $u$} as the set of questions for which the valid answers based on knowledge before and after the update differ: \vspace{-1mm}
\[
\mathcal{Q}(u)=
\left\{q:\mathcal{A}(q;K_\text{old})\neq\mathcal{A}(q;K_\text{new})\right\}.
\] %
We want to measure whether an agent's response reflects the updated knowledge $K_\text{new}$ when queried with probes in $\mathcal{Q}(u)$. We restrict probes to events that occur after the knowledge cutoff of the agent being evaluated, so that the agent's parametric knowledge does not contain $f_\text{new}$. At the same time, the agent always has access to $K_\text{new}$ through real-time search. We can therefore attribute a response that correctly reflects $K_\text{new}$ to successful agentic search. Our setup involves several key choices:

\textbf{\highlight{Design Decision 1:} Paired direct and indirect probes for evaluation.} Recall that we want to measure whether an agent will proactively search for consequential updates or events when $q$ does not explicitly mention them. However, we do not want to confound our analysis with the difficulty of finding the update itself. To separate the two, we define two types of complementary probes that operate on the same underlying update and compare performance on them. 

Let $\mathcal{Q}_\text{dir}(u)\subseteq\mathcal{Q}(u)$ be questions that explicitly ask which of $f_\text{old}$ or $f_\text{new}$ holds. We call these \textbf{\textit{direct probes}}. For example, ``\textit{Will the Bayeux Tapestry Museum be open to visitors?}'' in Figure~\ref{fig:probe_example}a is a direct probe. For these questions, no additional reasoning or inference on top of the updated fact $f_\text{new}$ is needed to answer the question correctly. Merely knowing the fact suffices. Therefore, these direct probes serve as a \textit{control} that allows us to isolate the difficulty of finding the update itself, given access to $K_\text{new}$. As they explicitly inquire about the update, answers to direct probes are straightforward to verify with an LLM judge.

We define \textbf{\textit{indirect probes}} as all probes for $u$ that are not direct probes: \vspace{-1mm}
\[
\mathcal{Q}_\text{ind}(u)=\mathcal{Q}(u)\setminus\mathcal{Q}_\text{dir}(u).
\]
The question \textit{``Suggest a five-day itinerary in Normandy''} in Figure~\ref{fig:probe_example}b is an example of an indirect probe. Although it never mentions the museum or asks about its status, the update changes its set of valid answers. An itinerary that includes the \textit{Bayeux Tapestry Museum} is no longer valid.
Our benchmark reports each agents' performance on both direct and indirect probes. %

There are two main challenges in constructing indirect probes. (i)~The question must not explicitly inquire about $f_\text{new}$ or the event, yet the update must still change the set of valid answers. (ii)~An agent may avoid an invalid answer by chance rather than by discovering the update. For example, it may omit the \textit{Bayeux Tapestry Museum} from an itinerary without ever learning of the closure. We must distinguish such cases from genuine discovery. Our design and selection of indirect probes address both challenges. 

\textbf{\highlight{Design Decision 2:} Instantiating indirect probes as recommendation requests.} Our indirect probes are recommendation requests for which the entity $x$ is a natural candidate before the update but is disqualified after it. Formally, let $\mathcal{C}(q;K)$ denote the set of entities eligible for the request $q$ under knowledge state $K$. 
Indirect probes, other than being in $\mathcal{Q}_\text{ind}(u)$, must satisfy: \vspace{-1mm}
\[
x\in\mathcal{C}(q;K_\text{old}) \,\,\, \text{and} \,\,\, x \notin \mathcal{C}(q;K_\text{new}). 
\] 
In other words, the update makes $x$ ineligible for the request. \citet{li-goyal-2025-memorization} introduce the term indirect probes for a specific construction, namely questions with list-style answers that include $x$ before the update but not after it. We instead give a formal definition of indirect probes as $\mathcal{Q}_\text{ind}(u)$ and adopt a similar construction to instantiate our benchmark.

Note that these recommendation-style indirect probes address challenge (i) above, but not challenge (ii) which we address next via design decision 3.

\textbf{\highlight{Design Decision 3:} Selecting indirect probes where the agent consistently recommends $\bm{x}$ before the update.}
Suppose the agent being evaluated consistently recommends $x$ for an indirect probe before the update, relying on $f_\text{old}$. We can then score their answers under $K_\text{new}$ (our actual evaluation setting) by the \textit{absence} of $x$ as a recommendation. %
E.g., if the agent consistently recommended \textit{Bayeux Tapestry Museum} before the closure, then its omission after the update can be reasonably attributed to having discovered the update. 

To determine whether an agent consistently recommends $x$ under $K_\text{old}$, we would need to block its access to every web page that reports the update. This is infeasible. 
However, we choose events that occur after the agent's knowledge cutoff, 
we therefore can sample responses from the agent \emph{with search disabled} and use them as a proxy for what it would have recommended before the update. 

For each candidate indirect probe, we sample multiple responses with search disabled and retain the probe only if the agent recommends $x$ in every response. This strict criterion 
addresses the (ii) concern, and allows us to attribute the omission of $x$ --- when search is enabled --- to discovering the update rather than to sampling variation.

\begin{table}[!h]
\centering
\caption{Examples from \sysname for its two types of updates, demises and ends of official tenures. The direct probe asks whether the person is still alive or still holds the position. The indirect probe never mentions the person, yet they would have been an obvious recommendation for the request before the update.}
\label{tab:examples}
\fontsize{8}{9.5}\selectfont
\renewcommand{\arraystretch}{1.15}
\begin{tabular}{@{}p{0.11\linewidth}p{0.86\linewidth}@{}}
\toprule
\multicolumn{2}{@{}l}{\textbf{Demises}} \\
\midrule
Update & Tony Hoare passed away on 5 March 2026. \\
Direct   & Is Tony Hoare, the British computer scientist, alive? \\
Indirect & I'm putting together an anthology on the history of programming language design, focusing on type safety and legacy design errors. Who are a few veteran British computer scientists who have been outspoken about early language design flaws that we could invite to write a new foreword for the book? \\
\cmidrule{1-2}
Update   & Nathalie Baye passed away on 17 April 2026. \\
Direct   & Is Nathalie Baye, the French actress, alive? \\
Indirect & We are assembling the jury for a new national grant supporting emerging French filmmakers. Who are some highly decorated veteran French actresses with multiple César wins who would have the gravitas to sit on the panel and review this year's final project submissions? \\
\midrule
\multicolumn{2}{@{}l}{\textbf{Tenures}} \\
\midrule
Update   & Anne Hidalgo left the position of mayor of Paris on 28 March 2026. \\
Direct   & Is Anne Hidalgo currently serving as mayor of Paris? \\
Indirect & We're organizing a global summit on the `15-minute city' concept and need municipal leaders of major European capitals to sign our new urban proximity pledge on behalf of their cities. Who are a few people we should consider reaching out to for this official sign-off? \\
\cmidrule{1-2}
Update   & Markwayne Mullin left the position of United States senator from Oklahoma on 23 March 2026. \\
Direct   & Is Markwayne Mullin currently serving as a United States senator from Oklahoma? \\
Indirect & Our sports advocacy group is looking to brief federal lawmakers who competed in collegiate wrestling about some upcoming athletic funding bills. Who makes sense to have in the mix for these briefings? \\
\bottomrule
\end{tabular}
\end{table}

\subsection{Automatic Benchmark Construction Pipeline}
\label{sec:construction}
\vspace{-3mm}

Each datapoint in our benchmark is a pair of direct and a complementary indirect probe that target the same underlying update (see Table~\ref{tab:examples} for examples). We next describe how we gather these updates and construct the probes automatically. Appendix~\ref{app:construction} gives further details on each step. 

\noindent \textbf{Gathering updates.} We source real-world events from Wikidata. We focus on two event types about people, namely demises and ends of official tenures (tenures for short). We choose these as  proofs of concept because both have enough support in Wikidata since frontier models' knowledge cutoffs, although our conceptual framework applies to other event types.

\noindent \textbf{Constructing indirect probes.}
For each update, we prompt a frontier LLM to generate direct and indirect probes. Design decision 3 retains an indirect probe for an agent only if the agent consistently recommends $x$ without search. We therefore first generate and filter indirect probes, and then generate direct probes only for the filtered subset. 
We provide the update and associated Wikipedia articles as context to the LLM and prompt it to propose five candidate indirect probes. Each probe asks for a shortlist that admits several plausible candidate entities without uniquely identifying the entity $x$ that the update is about. 
We also include category-specific constraints. For probes related to demises, the request must call for future participation. For tenures, the request must require authority that only the current holder of the position has. Across the five probes for an update, we vary who is making the request (e.g., a doctoral student), the scenario it arises in (e.g., seeking approval for a research visit), and which consequence of the update makes $x$ ineligible (only the department head can grant that approval, so if their tenure is over, they would be ineligible).

\noindent \textbf{Filtering indirect probes per agent.}
Following design decision 3, we sample three responses with search disabled for each agent and candidate indirect probe and retain the probe only if all three recommend $x$. An LLM judge decides whether each response recommends $x$. It is carefully prompted so that it only penalizes a response if it \textit{recommends} the disqualified entity, and does not penalize mere mentions. We filter probes separately for each agent, since agents differ in what they recommend. We filter at low reasoning effort, since what a model knows depends on its weights rather than its reasoning budget, and share the resulting subset across both effort settings so that their comparison uses the same questions.

\noindent \textbf{Final benchmark and scoring.} For each retained indirect probe, we prompt the LLM to generate a paired direct probe that explicitly asks which of $f_\text{old}$ and $f_\text{new}$ holds. An LLM judge scores a direct response as correct only if it explicitly states $f_\text{new}$, namely that the person has died (demises) or no longer holds the position (tenures). Responses that affirm $f_\text{old}$, express uncertainty, or evade the question are marked incorrect. For indirect probes, we reuse the judge from the filtering step.

\noindent \textbf{Implementation Details.}
We generate indirect probes with Gemini-3.1-Pro \citep{google2026gemini31pro} and direct probes with Gemini-3.7-Flash \citep{google2026gemini37flash}. We use Gemini-3.5-Flash-Lite \citep{google2026gemini35flashlite} as the judge for all indirect and direct responses, including those scored during filtering. Appendix~\ref{app:construction} lists the reasoning settings for each stage, and the full prompts are provided in the supplementary material. To validate the judge, one author labeled a stratified sample of 200 responses per category for each probe type, without access to the judge's labels. On indirect probes, the judge agreed with the human label on 199 (99.5\%) responses for demises and 197 (98.5\%) for tenures. On direct probes, it agreed on all 200 responses for tenures and on 198 (99.0\%) for demises. See Appendix~\ref{app:judge-validation} for more details.

\noindent \textbf{Benchmark Statistics.} We instantiate our benchmark with events dated between Feb and June 2026, yielding 2,078 demises and 751 tenures. The window begins in Feb because the latest reported cutoff among the models we evaluate is January 2026, that of Claude-Opus-4.7 (Table~\ref{tab:knowledge-cutoffs}). Generating five indirect probes per event gives 10,390 candidate probes for demises and 3,755 for tenures. Table~\ref{tab:filter-counts} reports the fraction of retained probes for each model after the filtering step described above.

\section{Experimental Setup}
\vspace{-3mm}
\textbf{Models Evaluated.}
We evaluate 8 frontier LLMs from different providers, covering both open- and closed-source models: Claude-Opus-4.7 \citep{anthropic2026opus47}, GPT-5.4 \citep{openai2026gpt54}, Gemini-3.1-Pro \citep{google2026gemini31pro}, Qwen3.7-Max \citep{qwen2026qwen37max}, DeepSeek-V4-Pro-0423 \citep{xu2026deepseek}, Kimi-K2.6 \cite{moonshot2026kimik26}, MiMo-V2.5-Pro \citep{xiaomi2026mimo25pro}, and GLM-5.1 \citep{zhipu2026glm51}. Providers expose reasoning effort in different ways, so we evaluate each model at two settings, chosen as the nearest available match to a low and a high effort budget. We refer to these as low and high throughout, and Table~\ref{tab:effort-map} lists the provider-specific values.

From each model's probes retained by the filtering step (Table~\ref{tab:filter-counts}) we sample 200 per category, drawing exactly one probe per event. Where fewer than 200 events survive the filter, we use all of them. This affects tenures for Claude-Opus-4.7, with 175 events, and MiMo-V2.5-Pro, with 115. All results in Table~\ref{tab:main-results} are computed on these model-specific samples.

\textbf{Native and shared harnesses for evaluation.}  Most providers ship a \textit{\native} harness with a first-party search tool, the provider's default system instructions, and other specifications. These reflect how users actually encounter these systems. However, these harnesses differ across providers and offer a limited view of the search trajectories, hindering both apples-to-apples comparison and systematic analysis. Therefore, we additionally evaluate every model under a common scaffold, which we call the \textit{\shared} harness. Models without a \textit{\native} harness are only evaluated using the \textit{\shared} harness.

We implement the \textit{\shared}  harness with Inspect AI \citep{inspect2024}. We fix the search backend and interaction protocol so that differences between models are not due to provider-specific search integrations. Each model receives the question and a common system prompt (Figure~\ref{fig:system-prompt}) that states the evaluation date, the model's knowledge cutoff, and instructions to search before answering potentially time-sensitive questions. The harness exposes a single web-search tool backed by Exa (\url{https://exa.ai}), which returns source titles, URLs, publication dates when available, and relevant text highlights. Models write their own queries and may search repeatedly before answering. We log every response, search call and tool output for post-hoc analysis.

\newcolumntype{C}{>{\centering\arraybackslash}p{2.2em}}
\newcommand{\miconheight}{0.9em}
\newcommand{\micon}[1]{\raisebox{-0.15em}{\includegraphics[height=\miconheight]{analysis/icons/#1.pdf}}}
\newcommand{\mname}[2]{\shortstack[l]{#1\\#2}}
\newcommand{\miname}[2]{\micon{#1}~#2}
\newcommand{\vsep}{\hspace{0.25em}{\color{black!30}\vline}\hspace{0.25em}}
\newcommand{\modelrule}{\arrayrulecolor{black!15}\midrule\arrayrulecolor{black}}
\newcommand{\dcg}[1]{\cellcolor{\ifdim #1pt<0pt green\else red\fi!\fpeval{min(abs(#1)*0.6,50)}!white}#1}

\begin{table}[t]
\centering
\caption{Direct (Dir.) and indirect (Ind.) probing accuracy (\%) for each model, harness and reasoning effort.
The \dropmark\ columns show the drop from direct to indirect accuracy. -- marks MiMo-V2.5-Pro (\shared, low), which kept calling tools without giving a final answer.\vspace{-3mm}
}
\label{tab:main-results}
\scriptsize
\setlength{\tabcolsep}{3pt}
\begin{tabular}{lc !{\vsep} CCC !{\vsep} CCC !{\vsep} CCC !{\vsep} CCC}
\toprule
 & & \multicolumn{6}{c!{\vsep}}{Demises} & \multicolumn{6}{c}{Tenures} \\
\cmidrule(lr){3-8}\cmidrule(lr){9-14}
 & & \multicolumn{3}{c}{Low} & \multicolumn{3}{!{\vsep}c!{\vsep}}{High}
   & \multicolumn{3}{c}{Low} & \multicolumn{3}{!{\vsep}c}{High} \\
\cmidrule(lr){3-5}\cmidrule(lr){6-8}\cmidrule(lr){9-11}\cmidrule(lr){12-14}
Model & Harness & Dir. & Ind. & \dropmark & Dir. & Ind. & \dropmark & Dir. & Ind. & \dropmark &  Dir.& Ind. & \dropmark \\
\midrule
 & \native     & 96.0 & 58.5 & \dcg{37.5} & 96.5 & 69.0 & \dcg{27.5} & 97.7 & 84.6 & \dcg{13.1} & 96.0 & 85.7 & \dcg{10.3} \\
\multirow{-2}{*}{\miname{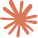}{Opus-4.7}}
 & \shared     & 96.5 & 26.5 & \dcg{70.0} & 97.0 & 60.0 & \dcg{37.0} & 96.0 & 90.9 & \dcg{5.1}  & 100.0 & 91.4 & \dcg{8.6}  \\
\modelrule
\multirow{2}{*}{\miname{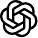}{5.4}}
 & \native     & 99.0 & 57.5 & \dcg{41.5} & 100.0 & 92.0 & \dcg{8.0}  & 86.0 & 76.5 & \dcg{9.5}  & 98.5 & 93.5 & \dcg{5.0}  \\
 & \shared     & 99.5 & 35.5 & \dcg{64.0} & 100.0 & 94.5 & \dcg{5.5}  & 96.5 & 83.5 & \dcg{13.0} & 99.5 & 94.0 & \dcg{5.5}  \\
\modelrule
 & \native     & 99.5 & 54.0 & \dcg{45.5} & 100.0 & 87.5 & \dcg{12.5} & 99.5 & 60.0 & \dcg{39.5} & 99.5 & 85.5 & \dcg{14.0} \\
\multirow{-2}{*}{\miname{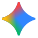}{3.1-Pro}}
 & \shared     & 99.5 & 44.0 & \dcg{55.5} & 99.5 & 75.5 & \dcg{24.0} & 99.0 & 74.5 & \dcg{24.5} & 100.0 & 83.5 & \dcg{16.5} \\
\modelrule
\multirow{2}{*}{\miname{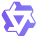}{3.7-Max}}
 & \native     & 21.0 & 30.0 & \dcg{-9.0} & 100.0 & 72.5 & \dcg{27.5} & 70.5 & 34.0 & \dcg{36.5} & 99.0 & 83.0 & \dcg{16.0} \\
 & \shared     & 73.5 & 31.0 & \dcg{42.5} & 100.0 & 77.5 & \dcg{22.5} & 95.5 & 48.5 & \dcg{47.0} & 100.0 & 86.0 & \dcg{14.0} \\
\modelrule
\miname{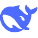}{V4-Pro} & \shared     & 97.5 & 77.5 & \dcg{20.0} & 100.0 & 84.0 & \dcg{16.0} & 92.5 & 91.5 & \dcg{1.0}  & 100.0 & 93.0 & \dcg{7.0}  \\
\modelrule
\miname{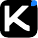}{K2.6}    & \shared     & 98.5 & 78.5 & \dcg{20.0} & 100.0 & 78.5 & \dcg{21.5} & 95.0 & 81.5 & \dcg{13.5} & 100.0 & 87.0 & \dcg{13.0} \\
\modelrule
 & \native     & 85.0 & 32.5 & \dcg{52.5} & 86.0 & 34.5 & \dcg{51.5} & 86.1 & 68.7 & \dcg{17.4} & 89.6 & 72.2 & \dcg{17.4} \\
\multirow{-2}{*}{\miname{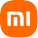}{V2.5-Pro}}
 & \shared     & 93.0 &  --  &           & 100.0 & 78.0 & \dcg{22.0} & 99.1 &  --  &           & 100.0 & 92.2 & \dcg{7.8}  \\
\modelrule
\multirow{2}{*}{\miname{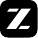}{5.1}}
 & \native     & 98.0 & 58.5 & \dcg{39.5} & 98.5 & 64.5 & \dcg{34.0} & 84.5 & 64.5 & \dcg{20.0} & 91.0 & 74.5 & \dcg{16.5} \\
 & \shared     & 94.5 & 66.5 & \dcg{28.0} & 100.0 & 86.5 & \dcg{13.5} & 99.5 & 92.5 & \dcg{7.0}  & 99.5 & 93.5 & \dcg{6.0}  \\
\bottomrule
\end{tabular}\vspace{-4mm}
\end{table}

\section{Main Results: The Gap Between Direct and Indirect Probing}
\vspace{-3mm}

Table~\ref{tab:main-results} reports results for all 8 models. We refer to a model, harness and reasoning effort together as a configuration. For each configuration, we report accuracy under direct and indirect probes in the Dir. and Ind. columns respectively. The \dropmark\ columns report the gap between the two.

\textbf{Indirect vs direct probing.} Indirect accuracy falls below direct accuracy in every configuration, with one exception (Qwen3.7-Max with the \native harness). On demises, the median gap is 42.5 points at low effort and 21.8 points at high effort with the \shared harness. With the \native harness that users actually encounter, the gap is large at 40.5 and 27.5 points respectively. Notably, Claude-Opus-4.7, a model at the frontier of other agentic search benchmarks, shows the largest gap at 70.0 points with the \shared harness at low effort. %
Overall, results in Table~\ref{tab:main-results} show that models that can retrieve an update when asked about it often fail to discover and account for the update when the request does not explicitly inquire about it.

\textbf{Demises vs tenures.} Interestingly, we find that the gap is typically larger on demises than on tenures (in 23/27 configurations). With the \native harness, the median gap on tenures is 18.7 points compared to 40.5 on demises at low effort, and 15.0 compared to 27.5 at high effort. The \shared harness shows the same pattern, with 13.0 compared to 42.5 points at low effort and 8.2 compared to 21.8 at high effort. Gemini-3.1-Pro and Qwen3.7-Max stand out at low effort, with tenure gaps of 39.5 and 36.5 points under the \native harness. We hypothesize that because officeholders change routinely, models are more inclined to check whether a person still holds a position than whether they are still alive. We examine this in Section~\ref{sec:failure-analysis}.

\textbf{High vs low reasoning effort.} Higher effort improves indirect accuracy on demises in 12/13 model-harness pairs, with a median gain of 31.5 points. However, the benefit is uneven. With the \shared harness, GPT-5.4 improves from 35.5\% to 94.5\%, while Kimi-K2.6 stays at 78.5\%. On tenures, indirect accuracy improves for all 13 pairs, but by a smaller median 9.0 points. We hypothesize that higher effort helps mainly by prompting models to check whether the recommended person is still alive or still holds the position. Models already do this more often for positions at low effort, which leaves less room to improve on tenures. Section~\ref{sec:failure-analysis} confirms this.

\providecommand{\miconheight}{0.9em}
\providecommand{\micon}[1]{\raisebox{-0.15em}{\includegraphics[height=\miconheight]{analysis/icons/#1.pdf}}}
\providecommand{\miname}[2]{\micon{#1}~#2}
\providecommand{\vsep}{\hspace{0.25em}{\color{black!30}\vline}\hspace{0.25em}}
\providecommand{\modelrule}{\arrayrulecolor{black!15}\midrule\arrayrulecolor{black}}
\begin{wraptable}{r}{0.45\textwidth}
\centering
\caption{Indirect probing accuracy (\%) on a common set of questions shared across all
models, run using the shared harness with both low and high reasoning effort. 
}
\label{tab:common-set}
\footnotesize
\setlength{\tabcolsep}{4pt}
\newcolumntype{E}{>{\centering\arraybackslash}p{2.2em}}
\begin{tabular}{l EE !{\vsep} EE}
\toprule
 & \multicolumn{2}{c!{\vsep}}{Demises \scriptsize{(n = 200)}} & \multicolumn{2}{c}{Tenures \scriptsize{(n = 172)}} \\
\cmidrule(lr){2-3}\cmidrule(lr){4-5}
Model & Low & High & Low & High \\
\midrule
\miname{gpt}{5.4}         & 26.0 & 90.5 & 76.7 & 89.5 \\
\modelrule
\miname{gemini}{3.1-Pro}  & 35.5 & 72.0 & 79.1 & 82.6 \\
\modelrule
\miname{deepseek}{V4-Pro} & 65.0 & 81.5 & 89.5 & 90.1 \\
\modelrule
\miname{kimi}{K2.6}       & 78.5 & 79.0 & 83.1 & 87.8 \\
\modelrule
\miname{glm}{5.1}         & 60.0 & 82.5 & 83.7 & 88.4 \\
\bottomrule
\end{tabular}
\vspace{-0.8em}
\end{wraptable}

\begin{figure}[t]
    \centering
    \includegraphics[width=0.9\linewidth]{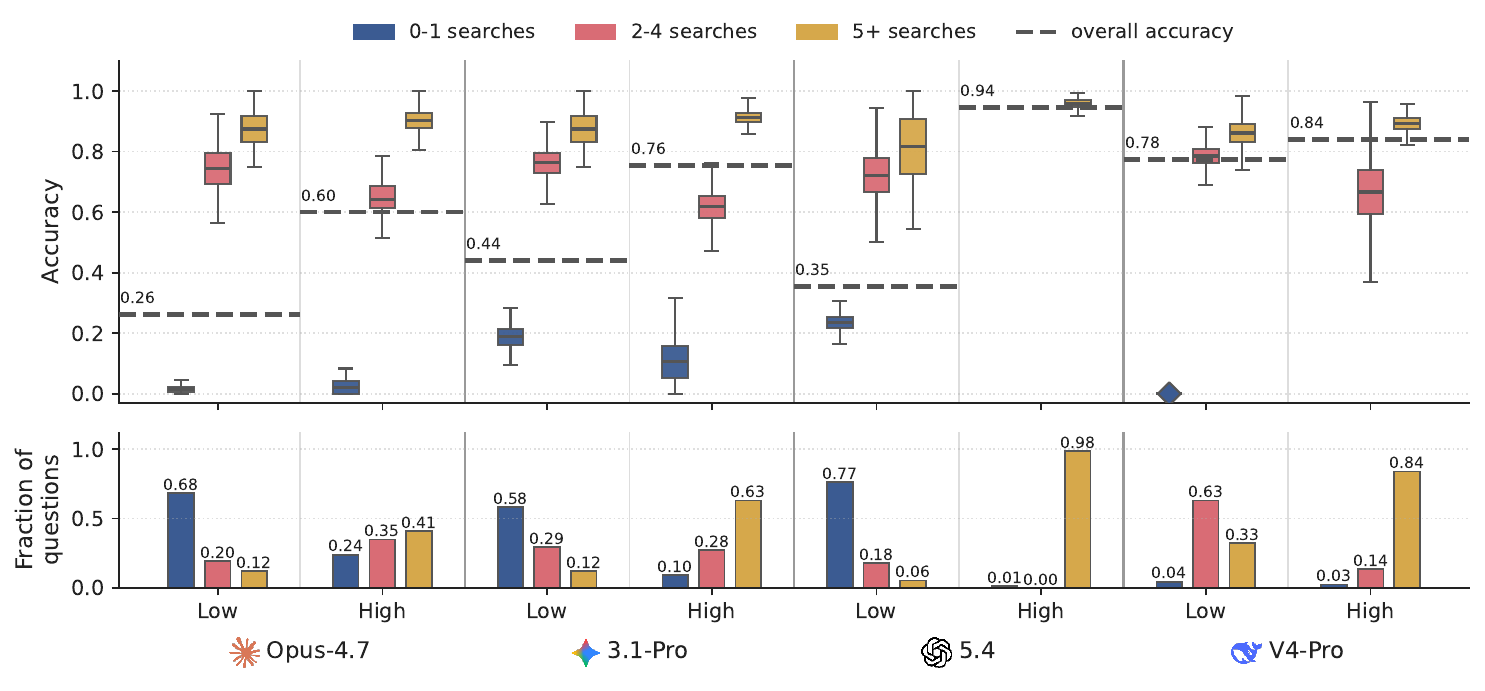}\vspace{-2mm}
    \caption{Across models, we observe comparable performance on indirect probes for demises using the \shared harness, when grouped by the number of search queries issued. Top, accuracy per group (with at least 9 probes) and its overall accuracy (dashed). Bottom, the share of probes in each group.}\vspace{-4mm}
    \label{fig:search-buckets}
\end{figure}

\textbf{Claude-Opus-4.7 rarely searches for demise probes.} Although Claude-Opus-4.7 performs strongly on tenures and on other popular agentic benchmarks \citep{anthropic2026opus47}, it has the largest gap on demises with the \shared harness at both effort levels (70.0 and 37.0 points). Figure~\ref{fig:search-buckets} groups the demise probes of Claude-Opus-4.7 and three other models by the number of search queries issued (see Appendix~\ref{app:search-buckets} for the remaining models). Among probes with at least two searches, Claude-Opus-4.7 answers 79\% correctly at low effort and 78\% at high effort, close to the median of 82\% across the other \shared configurations in Table~\ref{tab:main-results}. However, it issues no search on 68\% of demise probes at low effort and 23\% at high effort, while no other model exceeds 8\% at high effort. Its larger gap reflects how rarely it searches, not how well it uses what it finds. 

\textbf{Shared vs native harness.} Our analysis of search trajectories relies on the \shared harness, so models should perform comparably in both harnesses. At high effort they do, with indirect accuracy higher in the \shared harness for 9 of the 12 model-category pairs and at most 12.0 points lower for the rest. At low effort, the \shared harness trails on demises for Claude-Opus-4.7, GPT-5.4 and Gemini-3.1-Pro by up to 32.0 points, possibly because with reduced or no reasoning these models use a generic search tool less effectively than their provider's own. We therefore consider the \shared harness a reasonable setting in which to study search behavior.

\textbf{Comparing models.} To compare models on equal footing, we evaluate five models under the \shared harness on a common set of probes retained by all of them (Table~\ref{tab:common-set}). We include these five models to retain a sufficient number of probes in the intersection for a reliable comparison. At low effort, the three open-weights models are more accurate than the two closed ones on both categories. Search behavior accounts for much of this difference. GPT-5.4 and Gemini-3.1-Pro issue no search on 58.5\% and 56.5\% of demise probes, compared with at most 30.0\% for the open-weights models. When Gemini-3.1-Pro does search, it answers 80.5\% of probes correctly, in line with the open-weights models. On tenures, the closed models rarely skip search but search less, issuing a median of 3 queries against 7 to 11 for the open-weights models. At high effort, GPT-5.4 is the most accurate model on demises and performs comparably to the open-weights models on tenures. It issues a median of 16 queries on demise probes, more than any other model, and searches on 97.5\% of them. The spread across models is also larger on demises, at 52.5 points at low effort and 18.5 points at high effort, than on tenures, at 12.8 and 7.5 points. This is likely because models search more readily on tenures, skipping search less often than on demises in all configurations. 

\section{Error Analysis}
\label{sec:failure-analysis}
\vspace{-2mm}

To understand the gap between direct and indirect accuracy, we further categorize failures on indirect probes. We perform this analysis under the \shared harness as it logs every query, every search result and the context the model has before answering. We hypothesize that a model can fail at three stages: it never looks for the update, it looks for it but retrieval fails, or retrieves it and still gives an outdated recommendation. We convert these stages into four mutually exclusive failure modes. We assign each failure to the first one it satisfies, in the order below.

\begin{figure}[t]
    \centering
    \includegraphics[width=0.95\linewidth]{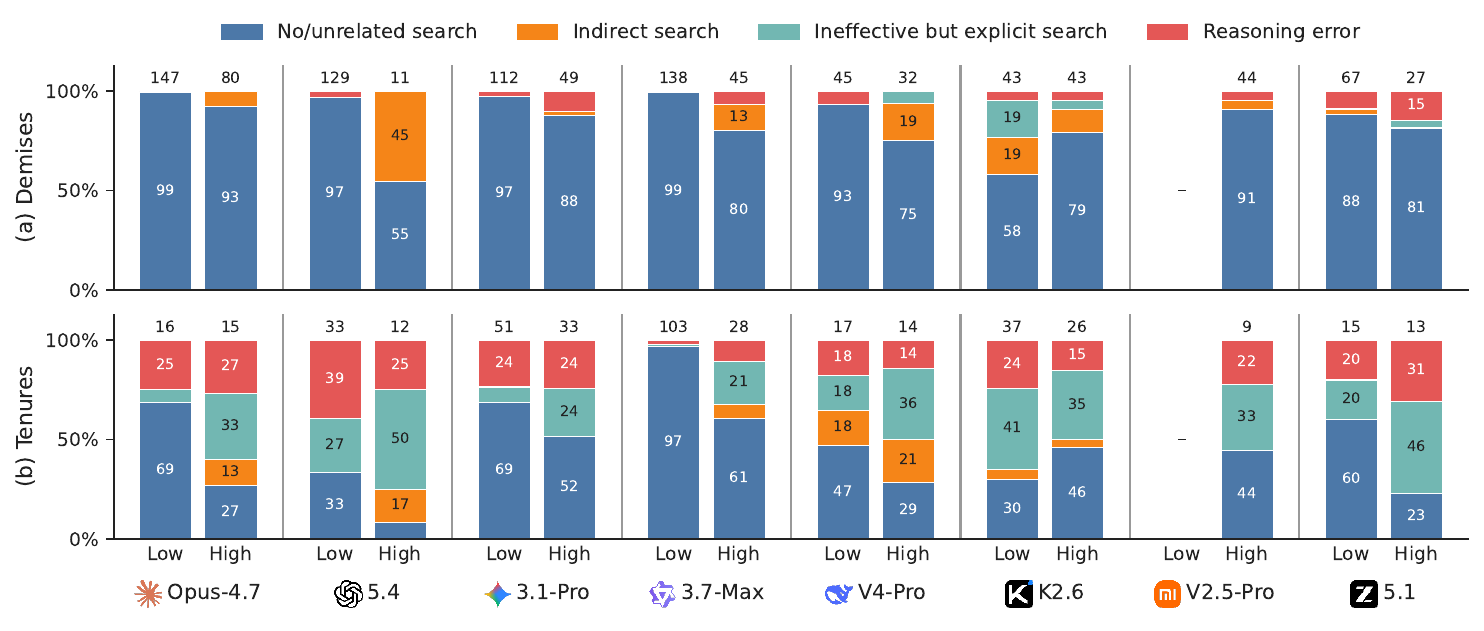}\vspace{-3mm}
    \caption{Composition of indirect-probe failure modes under the \shared harness for (a) demises and (b) tenures. Numbers above bars give the total number of failures.
    }\vspace{-3mm}
    \label{fig:failure-buckets}
\end{figure}

(1) \textbf{Reasoning error.} The model's context contains evidence of the update, yet its answer still makes the invalid recommendation. Prior work finds that models can favor parametric memory over conflicting context (Appendix~\ref{sec:related-knowledge-conflict}). 
(2) \textbf{Ineffective but explicit search.} The context contains no conclusive evidence of the update, but at least one query directly asks whether $f_\text{old}$ still holds, such as whether the person is alive or who currently holds the position. Such failures often reflect poorly phrased queries rather than limitations of the search tool.
(3) \textbf{Indirect search.} No query asks about the update explicitly, but at least one could reveal it indirectly, such as a search for the person's recent public appearances or actions in the role. Because our events are recent, such searches often return activity from before the update, which can suggest that $f_\text{old}$ still holds.
(4) \textbf{No/unrelated search.} No query could reveal the update, either because the model does not search or because its searches concern other parts of the request.

We assign failure categories to trajectories with Gemini-3.7-Flash at low reasoning effort (prompt is in the supplementary material). We verified the assigned mode for a stratified sample of 160 demise and 165 tenure failures, with an agreement of 98.1\% and 81.2\% respectively (Appendix~\ref{app:bucket-validation}). Most tenure disagreements involve searches whose relevance depends on background knowledge of the position, such as a search for the current members of a Senate committee, which would omit a senator who has left office. These mainly move failures between indirect and unrelated search, so our tenure conclusions rest on the two modes together. Appendix~\ref{app:failure-examples} shows representative failures.

\textbf{Not searching explicitly drives the gap, especially on demises.} On demises, failures with no/unrelated search make up at least 75\% of failures in 13 of the 15 configurations (Figure~\ref{fig:failure-buckets}a). On tenures, failures are spread more evenly across the four modes (Figure~\ref{fig:failure-buckets}b). Even so, indirect and no/unrelated search together account for at least 23\% of failures in every configuration and for a median of 50\% of failures. In 14 of the 15 configurations, failures with no/unrelated search account for all of the additional failures on demises compared with tenures. Since direct accuracy is similar on both categories, this explains the larger gap on demises.

\textbf{Higher reasoning helps mainly by getting models to search for the update.} We compare the share of probes in each failure mode at low and high effort for the seven models evaluated at both effort levels under the \shared harness. On demises, moving from low to high effort reduces failures with no/unrelated search by 28.9 points on average, nearly matching the average gain in indirect accuracy of 28.1 points. On tenures, the reduction is 9.1 points and the gain is 9.4 points. No other failure mode changes by more than 5.0 points for any model. Higher effort therefore improves accuracy almost entirely by getting models to search for the update on probes where they previously did not. The gain is larger on demises because failures with no/unrelated search are far more common there at low effort, which leaves more room to improve. Kimi-K2.6 is the exception and does not improve on demises because it already has few failures with no/unrelated search at low effort.

\section{Discussion}

\paragraph{Test-time compute versus performance.} Our work shows that agents struggle to keep up with a changing world because they rely on outdated assumptions. This happens when nothing in the wording of a request signals which underlying facts need to be checked. An obvious remedy is to instruct the agent to verify everything its answer depends on, for example by adding ``Before finalizing your answer, verify the facts it relies on'' to the prompt. This follows the common practice of spending test-time compute to improve performance \citep{snell2024scaling}. we speculate that it would help, since higher reasoning effort improves indirect accuracy by getting models to search for the update (\S\ref{sec:failure-analysis}). However, exhaustive verification scales poorly. The facts that matter are rarely stated in the answer itself. A response that recommends a speaker never explicitly states that the person is alive. The agent must therefore first enumerate what it has implicitly assumed, and then check each assumption for every candidate it considers. Every candidate that fails a check introduces a replacement that must be verified in turn.

One way to reduce this cost is to verify selectively.
Rather than verifying every assumption, an agent can prioritize those that are likely to have changed since its knowledge cutoff. Models already appear to follow this signal, as they check whether a person still holds a position more often than whether the person is alive. However, our results on demises show that how often a fact changes is not a sufficient criterion. A person's demise is a rare event, yet it invalidates every recommendation that involves them. A good verification strategy should therefore weigh both how likely an assumption is to be outdated and how much the answer depends on it. A promising direction is to train agents to decide what to verify, with the objective of balancing accuracy against test-time cost.

\paragraph{Continual Learning.} Even selective verification has a limit, since an agent can only check the assumptions it thinks of. In our benchmark we consider updates and their direct consequences, but real-world updates often have far reaching consequences. 
Consider a user who asks where to go on a safari next month. A drought around a popular wildlife reserve may drive the animals elsewhere. The reserve stays open but sightings may now be far poorer than at alternatives the agent never considers. An agent that verifies its assumptions would likely confirm that the reserve is open and recommend it, even though better options exist. Reasoning forward from a drought to sparse wildlife is easy, but reasoning backward from a safari to the many conditions that could spoil it is much harder. When the update that matters is several steps removed from the answer, verifying the answer's own assumptions is therefore not enough. Addressing such cases requires acquiring recent knowledge (in this case, knowing about the drought).

This motivates a complementary approach, which is to keep the agent's parametric memory up to date. 
Parametric memory is the source of the assumptions an agent starts from. An agent that has learned of a person's demise during training would not recommend them, and would not need to verify whether they are still alive.  
Interestingly, an update does not have to be recalled exactly to be useful, since search can supply the details. The agent only needs to recognize which of its assumptions are worth checking. For example, a model trained on reports of an upcoming election or of a person's failing health may be more inclined to check whether the officeholder has changed or whether the person is still alive, just as a model that knows of a drought can reason forward to poor sightings. Continual learning could therefore help not only by updating facts but also by indicating which facts may have changed. Additionally, updating the weights moves cost from test to training time, where it is paid once and amortized over all users and requests.

In practice, we expect the two approaches to be combined. No training schedule can keep a model fully current, leaving a gap between the most recent update and the date of a request. Changes that fall in this gap can only be caught through verification. Continual learning both shortens this gap, leaving fewer assumptions that could be outdated, and helps direct the remaining checks towards those most likely to have changed.

\section{Conclusion}
\vspace{-3mm}

 We introduce \sysname to track a new axis of agentic search capabilities: do AI agents proactively discover recent real-world changes that affect their responses? By pairing each indirect probe with a direct probe of the same change, we separate an agent's ability to find a change when asked from its ability to discover the change unprompted and account for its consequences. We show that frontier agents almost always find a change when asked directly, yet often miss it when the request does not inquire about it. Our error analysis traces many of these failures to agents never explicitly checking whether the relevant fact has changed, and shows that higher reasoning effort helps mainly by prompting such checks. Our findings highlight that benchmarks which state what needs to be found overestimate how well agents keep up with a changing world. %

\subsection*{AI use statement}

In this work, we used generative AI tools to generate the direct and indirect probes that make up our benchmark (\S\ref{sec:construction}), to judge model responses, and to support qualitative analysis by labeling failure modes in our error analysis (\S\ref{sec:failure-analysis}). We validated both the judge and the failure-mode labels against human annotations (Appendices~\ref{app:judge-validation} and~\ref{app:bucket-validation}). We also used AI coding assistants to help implement parts of our pipeline. We have not used generative AI tools to propose or refine hypotheses, design or provide feedback on our methodology or experiments, clean or reformat the dataset, interpret results, or develop our conceptual framework. Formulating mathematical claims, providing ingredients for or writing proofs, and translation are not applicable to this work.  Additionally, we used generative AI tools to edit the paper for readability. We have reviewed all AI-assisted work. We take responsibility for the final content of this work, including text, claims or artifacts produced with the aid of generative AI.

\subsubsection*{Acknowledgments}

AP was supported by a fellowship from Google and TG by NSF grant IIS2433072. DP is grateful to Schmidt Sciences, Google, and the Indian Institute of Science for supporting his group’s research.

\bibliography{iclr27}
\bibliographystyle{iclr2027_conference}

\clearpage
\appendix
\section{Extended Related Work}
\label{app:related-work}

\paragraph{Agentic Search Systems and Benchmarks.} Traditional Retrieval-Augmented Generation (RAG) systems supplement a model's parametric memory with a non-parametric retrieval corpus that is queried once using the input question \citep{lewis2020retrieval}. A line of work later made the retrieval decision part of generation itself. FLARE anticipates the upcoming sentence and, when that anticipation is low confidence, uses it to form a retrieval query \citep{jiang-etal-2023-active}. Self-RAG replaces this heuristic with learned reflection tokens, so the model decides at each step whether to retrieve and whether the retrieved evidence supports what it wrote \citep{asai2024self}. A parallel line of work gives the model a set of actions, now commonly called tool calls, rather than a single retrieve operation. WebGPT lets a model search, click, scroll, and quote inside a live browser \citep{nakano2021webgpt}, and ReAct interleaves reasoning with acting, so the model can revise its plan from what it observes \citep{yao2023react}. The agentic search systems we evaluate, often called agentic RAG, combine the two lineages \citep{li-etal-2025-search}. The model chooses when to invoke a search or browse tool, what query or URL to invoke it with, and when it has gathered enough evidence to answer.

Benchmarks for these systems have followed. GAIA poses general assistant tasks that require tool use, browsing, and multimodal reasoning \citep{mialon2024gaia}. BrowseComp narrows the setting to web research, with short-answer questions whose evidence is deliberately entangled and hard to locate \citep{wei2025browsecomp}. Because the live web changes between runs, BrowseComp-Plus rebuilds the same style of evaluation over a fixed human-verified corpus, which separates retrieval quality from reasoning over what was retrieved \citep{chen-etal-2026-browsecomp}. Other benchmarks shift the evaluation object from the final answer to the search process itself. DeepSearchQA scores comprehensiveness and reasoned stopping on exhaustive-list web tasks \citep{gupta2026deepsearchqa}, InfoDeepSeek scores the accuracy, utility, and compactness of the information an agent gathers from the live web \citep{xi2025infodeepseek}, and Mind2Web 2 judges long-horizon trajectories for both answer correctness and source attribution \citep{gou2025mind2web}. Two recent benchmarks ask how much of this performance is search rather than recall. LiveBrowseComp builds questions from low-salience facts published in the preceding ninety days, and shows that they largely defeat agents that score well on BrowseComp \citep{fan2026livebrowsecomp}. LiveNewsBench generates its questions automatically from clusters of news articles about a single recent event and refreshes them quarterly, so both the questions and answers change \citep{zhang2026livenewsbench}.

\paragraph{Temporal Freshness Benchmarks.} Facts differ in how long they stay true. MuLan separates Wikidata relations into mutable and immutable classes and finds that closed-book models are both less accurate and markedly less confident on the mutable ones \citep{fierro-etal-2024-mulan}. EverGreenQA labels questions rather than facts, and finds that this label predicts retrieval behavior \citep{pletenev-etal-2025-will}. RecencyQA also classifies how often an answer changes, from within an hour to never, and notes that this rate can itself depend on context \citep{piryani2026often}.

Benchmarks have taken up how well models handle questions whose answers change. RealTimeQA draws its questions from weekly news quizzes and releases a new set every week \citep{kasai2023realtime}. FreshQA instead fixes a set of 600 questions written by annotators to read like search queries, grouped by how quickly their answers turn over, and its maintainers re-verify the answers every week \citep{vu-etal-2024-freshllms}. Keeping either benchmark current is manual work, and its answers can fall out of date between updates. Later benchmarks instead construct questions from a source that records its own changes, so that answers can be refreshed automatically. PAT-Questions builds questions from templates over time-dependent Wikidata relations, and refreshes every answer by re-running the SPARQL query that produced it \citep{meem-etal-2024-pat}. EvolvingQA, GrowOVER, EvoWiki, and DynaQuest instead compare successive Wikipedia snapshots and generate questions from the passages or infobox fields that differ between versions \citep{kim-etal-2024-carpe,ko-etal-2024-growover,tang-etal-2025-evowiki,lin-etal-2025-dynaquest}. DailyQA follows the same recipe on a daily cycle, using revision logs to keep the answer key in step \citep{cheng2025dailyqa}.

\paragraph{Unstated Constraints and Agent-Initiated Information Seeking.} \citet{kaur-etal-2026-proper} represent the factors that determine task success as dimensions and separate the ones a user has stated from the ones they have not, treating the unstated ones as gaps an assistant should recognize on the user's behalf. Benchmarks that make such gaps discoverable through tool use find that recognizing them is the hard part. Implicit Intelligence places unstated constraints in environments where they can be found through a tool call, and finds that agents commonly act on the initial state without probing for the information that would reveal the constraint \citep{sirdeshmukh2026implicit}. DeepPlanning evaluates travel and shopping plans against rule-based checkers and reports that implicit environmental constraints such as limited seat availability or mismatched attraction opening hours are harder to detect than the requirements a user states \citep{zhang-etal-2026-deepplanning}. MapSatisfyBench recovers the decision factors a user left unspoken from behavior logs, and finds that answers are well grounded in the tool outputs an agent did collect while the tool that would supply the missing factor is rarely called \citep{bai2026mapsatisfybench}.

The same failure appears when the information need itself is underspecified. InteractComp builds queries whose intended target cannot be identified without more information, and reports that agents answer far fewer of them correctly than when the missing context is supplied, and that accuracy on them has not improved over a period in which BrowseComp scores rose sevenfold \citep{deng2026interactcomp}. DiscoBench separates noticing that a request is ambiguous from acting well once it is noticed, and finds that searching harder in place of resolving the ambiguity can do worse than guessing \citep{tao2026search}.

\paragraph{Knowledge Conflict and Downstream Behavior.}
\label{sec:related-knowledge-conflict}

A body of work studies what a model does when contextual evidence contradicts its parametric memory. \citet{longpre-etal-2021-entity} substitute the answer entity in gold passages, and find that models prefer the memorized answer over the one supplied in the context. \citet{du-etal-2024-context} find that the more familiar a model is with an entity, the less its answer is affected by contextual evidence about it. \citet{kortukov2024studying} replace synthetic conflicts with real corrective documents and find that updates fail less often than earlier work reported, but that a wrong parametric answer appearing in the context makes failure likelier. Conflicts also arise among the retrieved passages themselves. WikiContradict studies disagreements between Wikipedia passages, where both are equally credible, and finds that models rarely convey that the evidence is inconsistent \citep{hou2024wikicontradict}.

Knowledge editing writes the update into the weights, and MQuAKE and RippleEdits find that editing methods recall the edited fact but leave its consequences unchanged \citep{zhong-etal-2023-mquake,cohen-etal-2024-evaluating}. KUP separates direct probes that ask for the updated fact from indirect probes whose answer changes once the update holds, and finds that models memorize the update and still fail the indirect version \citep{li-goyal-2025-memorization}. TRACK supplies the updates in context instead, and finds that providing the correct fact can leave a model worse off than providing nothing, with the degradation growing as more conflicts enter a single reasoning chain \citep{feng-etal-2026-tracking}. IMPRINTBENCH compares both update families on the same corpus and scores them on six capabilities, including implicit relevance, where a query invokes an update through its timing or topic rather than naming it, and finds that no method succeeds across all six \citep{kalra2026measuring}.

A separate line of work asks what changes when the fact is temporal rather than counterfactual. HoH builds retrieval corpora in which outdated and current documents about the same fact coexist, and shows that the outdated ones degrade answers even when the current ones are retrieved alongside them \citep{ouyang-etal-2025-hoh}. WikiRecentChanges pairs stable and recently changed Wikidata facts with contextual statements that conflict with the model, and finds that models reason about change for the facts that did change and almost never for the stable ones, yet this reasoning rarely reaches the final answer \citep{wallat-etal-2026-facts}. STALE and HorizonBench build interaction histories in which a later event implicitly invalidates an earlier belief without restating it, and both find that systems identify the current state and still act on the old one \citep{chao2026stale, li2026horizonbench}.

In all of this work the update is already available to the model, either in its context or its weights, so the only research question is whether the response changes.  In our setting the update is not provided, so the agent must first suspect that what it recalls may be outdated and then search for the change. Of our failure categories, only reasoning errors correspond to the failure this line of work studies.

\section{Benchmark construction details}
\label{app:construction}

This appendix gives further details on each step of the pipeline described in \S\ref{sec:construction}. Table~\ref{tab:construction-settings} lists the model and reasoning effort used at each stage. The full prompts for every stage are provided in the supplementary material.

\begin{table}[t]
\centering
\caption{Models and reasoning efforts used at each stage of the \sysname
construction pipeline.}
\label{tab:construction-settings}
\small
\setlength{\tabcolsep}{6pt}
\begin{tabular}{lll}
\toprule
Stage & Model & Reasoning effort \\
\midrule
Indirect probe generation & Gemini-3.1-Pro &  High \\
Direct probe generation   & Gemini-3.7-Flash &  Low \\
Response judging          & Gemini-3.5-Flash-Lite &  Low \\
\bottomrule
\end{tabular}
\end{table}

\noindent \textbf{Gathering updates.} For each event, we record the affected entity $x$ and the corresponding update $u=(f_\text{old},f_\text{new})$. For demises, $f_\text{old}$ states that $x$ is alive and $f_\text{new}$ states that $x$ is deceased. For tenures, $f_\text{old}$ states that $x$ holds a particular position and $f_\text{new}$ states that $x$ no longer holds it. Both event types revoke $x$'s eligibility for a range of recommendation requests, which enables automated generation of many distinct probes per update. We require each event to occur after the most recent reported knowledge cutoff among the models we evaluate. An agent that already knows about the event could answer the probe from its memory, and the probe would then measure recall rather than discovery through search. We retain an update only if $x$ has a linked English Wikipedia article, which favors people that agents are likely to know well enough to recommend without search, as the filtering step requires. For tenures, we additionally require that $x$ does not currently hold the position, which removes cases of reelection where $f_\text{old}$ holds again.

\noindent \textbf{Constructing indirect probes.} We retrieve the full text of $x$'s English Wikipedia article and, for tenures, the position's Wikidata description and English Wikipedia article when available. We give this context and the update to the LLM and ask for natural, present-tense recommendation requests $q$ that do not name $x$ and satisfy $x\in\mathcal{C}(q;K_\text{old})$ and $x\notin\mathcal{C}(q;K_\text{new})$. The LLM uses the biographical and position context to generate requests for which $x$ is an obvious candidate before the update.

\noindent \textbf{Judging responses.} We judge whether a response to an indirect probe is consistent with $u$ using an LLM judge with a prompt specialized to each event type. Given the update $u$, indirect probe $q$, and agent response $a$, the judge assigns a binary label $J(u,q,a)\in\{0,1\}$. It assigns $0$ if $a$ recommends $x$ in a way that relies on $f_\text{old}$ and is therefore invalid under $K_\text{new}$, and $1$ otherwise. A response can receive $1$ by excluding $x$ or mentioning $x$ only in ways consistent with $K_\text{new}$. It need not explicitly state the update. For demises, a response receives $0$ if it treats $x$ as available for future participation. Historical references or mentions that acknowledge the death without recommending such participation are accepted. For tenures, a response receives $0$ if it presents $x$ as currently holding the ended position or exercising authority conferred by it. Historical references and recommendations based on independent qualifications are accepted provided they do not imply continued occupancy of that position. This judgment measures consistency with the targeted update, rather than the overall quality or correctness of the response. We use the same judge during filtering and evaluation.

\noindent \textbf{Filtering indirect probes per agent.} For each candidate probe $q$, we sample three responses $a_1,a_2,a_3$ from the agent with search disabled and retain the probe if and only if \[ J(u,q,a_j)=0 \qquad\text{for every }j\in\{1,2,3\}. \] The agent's no-search accuracy on its retained probes is therefore zero by construction.

\noindent \textbf{Final benchmark and scoring.} We generate each direct probe by prompting an LLM to fill a category-specific template using the context gathered for $x$. For demises, the template is ``Is [name], the [short identifying description], alive?'', where the description uses stable biographical attributes to disambiguate $x$. For tenures, it is ``Is [name] currently serving as [position]?'', with the position qualified by an organization, institution, or jurisdiction when necessary.

\section{Judge validation}
\label{app:judge-validation}

We validate the judge on responses produced at evaluation time with search enabled. For each category and probe type, we sample 100 responses the judge labeled correct and 100 it labeled incorrect. Responses are drawn evenly across configurations in Table~\ref{tab:main-results} for each judge label. For indirect probes, most responses the judge accepts still mention the affected person, typically alongside a note about the update, and these are the hardest to judge. We therefore draw 85 of the 100 accepted responses from those that name the person and 15 from those that do not.

One author labeled every sampled response using the same criteria given to the judge, blind to the judge's label and the configuration.  Table~\ref{tab:judge-confusion} compares the judge's labels with the human labels. Disagreements are rare and run in both directions.

\begin{table}[t]
\centering
\caption{Confusion matrices comparing the judge's labels with human labels. For each probe type and category we sampled 100 responses the judge labeled correct and 100 it labeled incorrect.}
\label{tab:judge-confusion}
\small
\setlength{\tabcolsep}{6pt}
\begin{tabular}{ll cc cc}
\toprule
 & & \multicolumn{2}{c}{Demises} & \multicolumn{2}{c}{Tenures} \\
\cmidrule(lr){3-4}\cmidrule(lr){5-6}
 & & \multicolumn{2}{c}{Human label} & \multicolumn{2}{c}{Human label} \\
\cmidrule(lr){3-4}\cmidrule(lr){5-6}
Probe & Judge label & Correct & Incorrect & Correct & Incorrect \\
\midrule
\multirow{2}{*}{Indirect} & Correct   & 99 & 1   & 99  & 1   \\
                          & Incorrect & 0  & 100 & 2   & 98  \\
\midrule
\multirow{2}{*}{Direct}   & Correct   & 100 & 0  & 100 & 0   \\
                          & Incorrect & 2   & 98 & 0   & 100 \\
\bottomrule
\end{tabular}
\end{table}

\section{Model knowledge cutoffs}
\label{app:cutoffs}

Table~\ref{tab:knowledge-cutoffs} lists, for each model we evaluate, the knowledge cutoff its provider reports and the source we take it from. Four of the eight providers report no cutoff.

\begin{table}[t]
\centering
\caption{Reported knowledge cutoff of each evaluated model, as stated by its
provider. We resolve each cutoff to the last day of the reported month and
substitute it for \texttt{\{knowledge\_cutoff\}} in the \shared harness system
prompt (Figure~\ref{fig:system-prompt}) in \texttt{YYYY-MM-DD} form. For models whose provider reports no cutoff we substitute January 2026, the latest reported cutoff among the models we evaluate. Every event in \sysname postdates all of these dates, so the instruction to search for information after the cutoff applies to every probe.}
\label{tab:knowledge-cutoffs}
\small
\setlength{\tabcolsep}{6pt}
\begin{tabular}{lll}
\toprule
Model & Reported knowledge cutoff & Source \\
\midrule
Claude-Opus-4.7 & January 2026 & \citep{anthropic2026transparency} \\
GPT-5.4         & August 31, 2025 & \citep{openai2026gpt54docs} \\
Gemini-3.1-Pro  & January 2025 & \citep{google2026gemini3docs} \\
Qwen3.7-Max     & Not reported & -- \\
DeepSeek-V4-Pro & Not reported & -- \\
Kimi-K2.6       & Not reported & -- \\
MiMo-V2.5-Pro   & December 2024 & \citep{xiaomi2026mimodocs} \\
GLM-5.1         & Not reported & -- \\
\bottomrule
\end{tabular}
\end{table}

\section{Benchmark statistics}
\label{app:filter-counts}

Table~\ref{tab:filter-counts} reports the number of events and generated
probes in each category, and the number of probes each model retains after the
pre-event filter.

\begin{table}[t]
\centering
\caption{Number of indirect probes per category before and after the pre-event
filter. Generated counts are shared across models. Retained counts are the
probes for which all three no-search responses from that model include the
affected person.}
\label{tab:filter-counts}
\small
\setlength{\tabcolsep}{6pt}
\begin{tabular}{lrr}
\toprule
 & Demises & Tenures \\
\midrule
Events            & 2,078  & 751   \\
Generated probes  & 10,390 & 3,755 \\
\midrule
\multicolumn{3}{l}{\textit{Retained probes after filtering}} \\
Claude-Opus-4.7   &  935   &  299   \\
GPT-5.4           & 1,304 & 766 \\
Gemini-3.1-Pro    & 3,092 & 1,712 \\
Qwen3.7-Max       & 566   & 353 \\
DeepSeek-V4-Pro   & 749   & 589 \\
Kimi-K2.6         & 781   & 397 \\
MiMo-V2.5-Pro     & 530   & 185 \\
GLM-5.1           & 726   & 453 \\
\bottomrule
\end{tabular}
\end{table}

\section{Reasoning effort settings}
\label{app:effort-map}

Table~\ref{tab:effort-map} lists the provider-specific effort value behind each
of the low and high reasoning settings reported in the main results.

\begin{table}[t]
\centering
\caption{Mapping from the low and high reasoning settings used throughout the
paper to the provider-specific effort value passed to each model.}
\label{tab:effort-map}
\small
\setlength{\tabcolsep}{6pt}
\begin{tabular}{lll}
\toprule
Model & Low & High \\
\midrule
Claude-Opus-4.7 & Disabled & High \\
GPT-5.4         & None & High \\
Gemini-3.1-Pro  & Low & High \\
Qwen3.7-Max     & Disabled & Enabled \\
DeepSeek-V4-Pro & Disabled & High \\
Kimi-K2.6       & Disabled & Enabled \\
MiMo-V2.5-Pro   & Disabled & Enabled \\
GLM-5.1         & Disabled & Enabled \\
\bottomrule
\end{tabular}
\end{table}

\section{Shared harness system prompt}
\label{app:system-prompt}

Figure~\ref{fig:system-prompt} reproduces the system prompt used in the
\shared harness. The placeholders are filled per run, with
\texttt{\{current\_date\}} set to the date the evaluation was run and
\texttt{\{knowledge\_cutoff\}} set to the cutoff listed for that model in Table~\ref{tab:knowledge-cutoffs}.

\begin{figure}[t]
\centering
\begin{lstlisting}[
  basicstyle=\small\ttfamily,
  frame=single,
  breaklines=true,
  breakindent=0pt,
  columns=fullflexible,
  xleftmargin=1em,
  xrightmargin=1em,
  framexleftmargin=0.5em,
  framexrightmargin=0.5em,
]
Current date: {current_date}. Model knowledge cutoff: {knowledge_cutoff}. Interpret 'current', 'currently', 'now',
  'today', 'recently', 'this year', 'latest', and similar terms relative to the current date. If a fact may have
  changed after the model knowledge cutoff, or is otherwise time-sensitive, call the web_search tool before
  answering.
\end{lstlisting}
\caption{System prompt supplied to every model in the \shared harness.}
\label{fig:system-prompt}
\end{figure}

\section{Accuracy by number of searches for the remaining models}
\label{app:search-buckets}

\begin{figure}[h]
    \centering
    \includegraphics[width=0.9\linewidth]{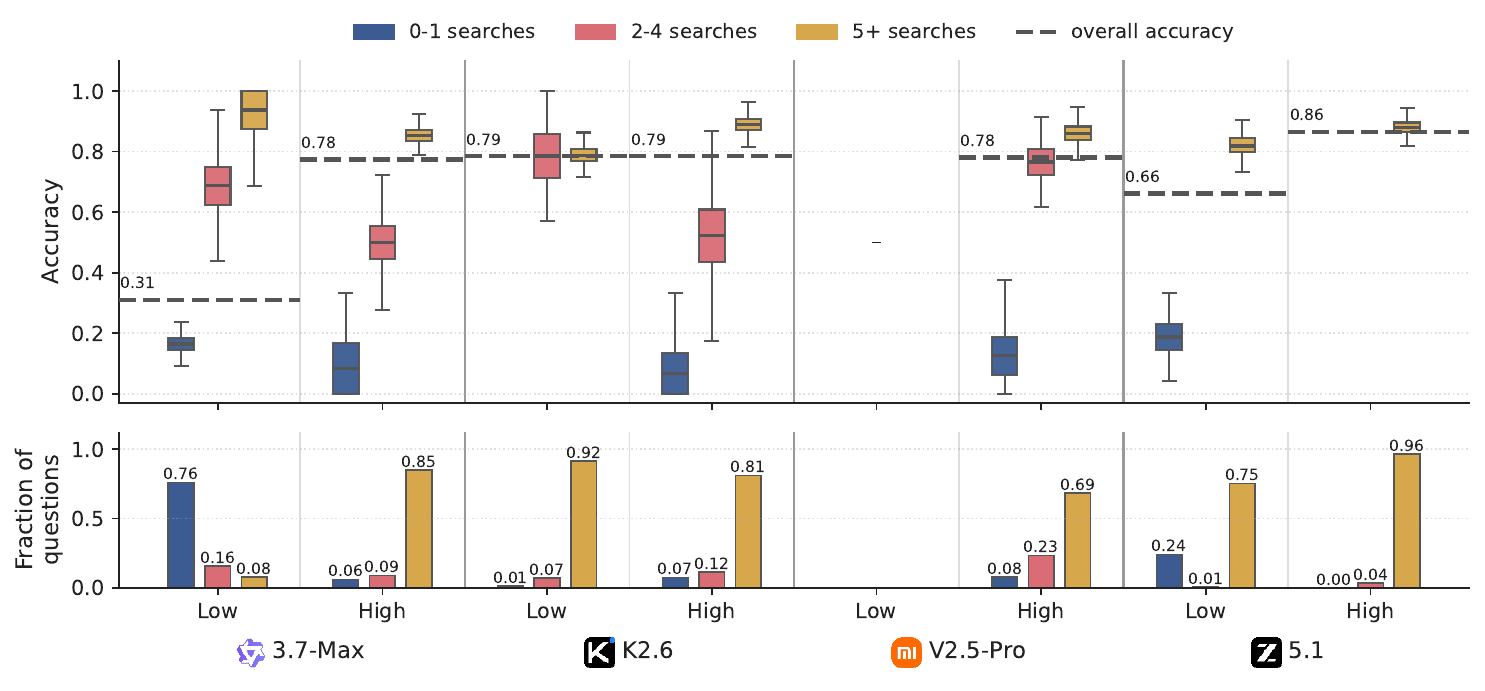}
    \caption{Indirect probe accuracy on demises under the \shared harness for the models not shown in Figure~\ref{fig:search-buckets}, grouped by the number of search queries issued. Top, accuracy per group (with at least 9 probes) and its overall accuracy (dashed). Bottom, the share of probes in each group.}
    \label{fig:search-buckets-remaining}
\end{figure}

Figure~\ref{fig:search-buckets} groups the demise probes of four models under the \shared harness by the number of search queries issued. Figure~\ref{fig:search-buckets-remaining} shows the same grouping for the remaining four models, Qwen3.7-Max, Kimi-K2.6, MiMo-V2.5-Pro and GLM-5.1. MiMo-V2.5-Pro has no low effort result, since it kept calling tools without giving a final answer (Table~\ref{tab:main-results}).

\section{Failure mode validation}
\label{app:bucket-validation}

We validate the failure modes of \S\ref{sec:failure-analysis} for both categories, using failures under \shared search in the same 15 configurations. A failure with no search query is assigned no/unrelated search by construction, so we sample only failures that issued at least one query. We take every failure in the smaller failure modes and sample the larger ones evenly across configurations, which gives 160 demise and 165 tenure failures.

One author checked each assigned failure mode against the model's queries and the context it had before its final answer, using the criteria given to the classifier. The author was shown the classifier's evidence excerpt and query labels. Table~\ref{tab:bucket-confusion} compares the assigned failure modes with the verified ones. The author agreed with 157 of 160 demise failures (98.1\%, Cohen's $\kappa = 0.97$) and 134 of 165 tenure failures (81.2\%, $\kappa = 0.74$).

On tenures, most disagreements concern searches that could reveal the change without asking about the person or the position itself. For example, when recommending a senator, a model may search for the current members of the Senate Judiciary Committee. The search asks about the committee rather than the senator's seat, but a senator who has left office would be missing from the list. Recognizing that such a search could reveal the change requires background knowledge about the position. Disagreements of this kind mainly move failures between indirect and unrelated search. Our conclusions are robust to this.

\begin{table}[t]
\centering
\caption{Confusion matrices comparing the failure mode assigned by the classifier with the failure mode verified by an author. No/unrelated search rows include only failures that issued at least one search query.}
\label{tab:bucket-confusion}
\small
\setlength{\tabcolsep}{5pt}
\begin{tabular}{ll cccc}
\toprule
 & & \multicolumn{4}{c}{Verified failure mode} \\
\cmidrule(lr){3-6}
Category & Assigned failure mode & Reasoning & Explicit & Indirect & No/unrelated \\
\midrule
\multirow{4}{*}{Demises} & Reasoning error                   & 33 & 0  & 1  & 1  \\
                              & Ineffective but explicit search   & 1  & 12 & 0  & 0  \\
                              & Indirect search                   & 0  & 0  & 42 & 0  \\
                              & No/unrelated search               & 0  & 0  & 0  & 70 \\
\midrule
\multirow{4}{*}{Tenures}       & Reasoning error                   & 43 & 1  & 1  & 0  \\
                              & Ineffective but explicit search   & 1  & 36 & 4  & 4  \\
                              & Indirect search                   & 0  & 1  & 9  & 5  \\
                              & No/unrelated search               & 2  & 3  & 9  & 46 \\
\bottomrule
\end{tabular}
\end{table}

\section{Representative failure cases}
\label{app:failure-examples}

We show a representative failure for each failure mode under \shared search. Each case gives the probe, the change the model should have accounted for, the model's answer to the direct probe for the same change in the same configuration, the search queries it issued, the key evidence in its context before the final answer and an excerpt of its answer. Excerpts are shortened with [\dots].

\newtcolorbox{failurecase}[1]{
  enhanced, breakable,
  colback=white, colframe=black!55, colbacktitle=black!8, coltitle=black,
  boxrule=0.5pt, arc=1pt, left=5pt, right=5pt, top=4pt, bottom=4pt,
  fonttitle=\small\bfseries, fontupper=\small,
  title={#1},
}
\newcommand{\failfield}[2]{\par\noindent\makebox[5.5em][l]{\textbf{#1}}\parbox[t]{\dimexpr\linewidth-5.5em\relax}{\raggedright #2}\par\smallskip}
\newcommand{\failquery}[2]{\par\hangindent=1.4em\hangafter=1\makebox[1.4em][l]{(#1)}\texttt{#2}}

\subsection{Demises}

\begin{failurecase}{Reasoning error \quad\textmd{GLM-5.1, high effort}}
\failfield{Probe}{Our regional sports museum is putting together a special interactive exhibit highlighting Bay Area football history. [\dots] We're ideally looking for legendary former Cal quarterbacks who have been inducted into the San Jose Sports Hall of Fame and went on to have long professional careers. Who would be some strong candidates to lead the selection process?}
\failfield{Change}{Craig Morton passed away on 9 May 2026.}
\failfield{Queries}{\failquery{1}{San Jose Sports Hall of Fame inductees list}
\failquery{2}{legendary Cal Berkeley quarterbacks professional career}
\failquery{3}{Steve Bartkowski Cal Berkeley quarterback San Jose Sports Hall of Fame}
\failquery{4}{Craig Morton Cal quarterback San Jose Sports Hall of Fame}}
\failfield{Evidence}{Result for query (4), Sports Illustrated, published 12 May 2026. \newline
\emph{``Cal Quarterback Great Craig Morton Dies at 83 [\dots] Craig Morton, one of Cal's greatest quarterbacks and a two-time starter in the Super Bowl during his 18-year NFL career, died Saturday.''}}
\failfield{Answer}{\emph{``Based on my research, there are two standout candidates who meet all your criteria [\dots] 1. Craig Morton. San Jose Sports Hall of Fame, Class of 2002 [\dots] Both men are strong candidates, but Craig Morton may have a slight edge for your purposes [\dots]''}}
\end{failurecase}

\begin{failurecase}{Ineffective but explicit search \quad\textmd{DeepSeek-V4-Pro, high effort}}
\failfield{Probe}{I'm organizing a vintage automotive showcase focused on mid-century flathead V8s and early slingshot dragsters. We need some guest-curators with hands-on experience modifying these specific engines back in the day to select the best cars for our main lineup. Who makes sense to have in the mix?}
\failfield{Change}{Ed Iskenderian passed away on 4 February 2026.}
\failfield{Direct}{Correct on \emph{``Is Ed Iskenderian, the American hot rodder and businessman, alive?''}}
\failfield{Queries}{\failquery{1}{Ed Iskenderian alive 2025 hot rod cams flathead}
\failquery{2}{Alex Xydias So-Cal Speed Shop alive 2025}
\failquery{3}{"TV Tommy" Ivo alive 2025 dragster}
\failquery{4}{Ed Pink engine builder alive 2025 flathead}
\failquery{5}{Don Garlits alive 2025}}
\failfield{Evidence}{None. All five results for the explicit query (1) predate the death and describe his 103rd and 104th birthday celebrations, for example \emph{``Ed Iskenderian turns 104: The Camfather honored with a celebration''} (NHRA, 2025) and \emph{``Celebrating Ed Iskenderian's 104th Birthday with the Man Himself''} (Hagerty, 10 September 2025).}
\failfield{Answer}{\emph{``[\dots] I verified current statuses before answering, so here's who actually makes sense to invite. Living legends you can actually book as guest curators. 1. Ed `Isky' Iskenderian, the priority pick for the flathead side. Now 105 (turned 104 in July 2025 and still active), he's arguably the last living legend of the flathead speed-equipment era. [\dots]''}}
\end{failurecase}

\begin{failurecase}{Indirect search \quad\textmd{DeepSeek-V4-Pro, high effort}}
\failfield{Probe}{I'm putting together a retrospective on the legacy of the New German Cinema. To kick things off, I want to invite a veteran filmmaker from that era who later expanded into independent television and experimental broadcasting to guest-curate a special lineup of counter-cinema TV slots for our upcoming festival. Who are a few people we should consider reaching out to for this programming role?}
\failfield{Change}{Alexander Kluge passed away on 25 March 2026.}
\failfield{Direct}{Correct on \emph{``Is Alexander Kluge, the German author and film director, alive?''}}
\failfield{Queries}{\failquery{1}{Alexander Kluge dctp independent television still active 2025}
\failquery{2}{New German Cinema filmmakers independent television experimental broadcasting dctp}
\failquery{3}{Edgar Reitz filmmaker still alive 2025 2026}}
\failfield{Evidence}{None. The indirect query (1) asks whether Kluge is still active rather than alive, and its results show his production company dctp still operating, with the most recent dated result a new dctp podcast announced on 4 June 2025. The model asks whether Edgar Reitz is still alive in query (3) but never asks this about Kluge.}
\failfield{Answer}{\emph{``[\dots] the person who most perfectly matches your description is not only alive but still extraordinarily active. [\dots] If you can only reach out to one person, make it Alexander Kluge (b. 1932). [\dots] Still active: dctp continues today as a digital platform (dctp.tv) with new collections, podcasts, and collaborations. [\dots]''}}
\end{failurecase}

\begin{failurecase}{No/unrelated search \quad\textmd{Qwen3.7-Max, high effort}}
\failfield{Probe}{I'm putting together a gallery show about integrating mobile tech into traditional en plein air landscapes. We want to commission a brand new, large-scale digital piece from an established master. Who are a few veteran British artists famous for embracing iPad drawing and digital tools in their landscapes that we should try to commission?}
\failfield{Change}{David Hockney passed away on 11 June 2026.}
\failfield{Direct}{Correct on \emph{``Is David Hockney, the British painter and photographer, alive?''}}
\failfield{Queries}{\failquery{1}{veteran British artists iPad drawing digital landscapes en plein air}
\failquery{2}{established British artists famous iPad art landscapes}
\failquery{3}{David Hockney iPad landscapes en plein air Yorkshire Wolds}
\failquery{4}{famous British landscape artists using iPad digital art established veteran}}
\failfield{Evidence}{None. Query (3) names Hockney but asks only about his iPad landscapes. Its results describe these works, with the most recent dated results covering the October 2025 Sotheby's sale of his Yorkshire Wolds iPad drawings.}
\failfield{Answer}{\emph{``Here's the honest answer: this question really has one towering, obvious answer, and then a longer tail of less-famous practitioners. [\dots] The Undisputed Master: David Hockney (b. 1937). Hockney is the answer to your question and is essentially in a category of his own. [\dots]''}}
\end{failurecase}

\subsection{Tenures}

\begin{failurecase}{Reasoning error \quad\textmd{Qwen3.7-Max, high effort}}
\failfield{Probe}{I'm putting together an official intergovernmental working group on European border security. We need to include European heads of government who have a track record of building physical border barriers and who originally trained as lawyers. Who are a few leaders we should consider reaching out to?}
\failfield{Change}{Viktor Orbán left the position of Prime Minister of Hungary on 9 May 2026.}
\failfield{Direct}{Correct on \emph{``Is Viktor Orbán currently serving as Prime Minister of Hungary?''}}
\failfield{Queries}{\failquery{1}{European heads of government lawyers built border fence barrier}
\failquery{2}{Viktor Orbán law degree border fence Hungary}
\failquery{3}{Evika Siliņa Latvia education law degree border fence Belarus}
\failquery{4}{Karl Nehammer Austria education law degree border fence}
\failquery{5}{Miro Cerar Slovenia law degree border fence Croatia}
\failquery{6}{Robert Fico Slovakia law degree border fence barrier}}
\failfield{Evidence}{Result for query (2), Wikipedia. \newline
\emph{``Viktor Mihály Orbán (born 31 May 1963) is a Hungarian lawyer and politician who served as the prime minister of Hungary from 1998 to 2002 and from 2010 to 2026.''}}
\failfield{Answer}{\emph{``Based on your criteria, here are European heads of government who both trained as lawyers and have track records of building physical border barriers. Viktor Orbán (Hungary) [\dots] Current status: Prime Minister since 2010 (previously 1998-2002) [\dots]''}}
\end{failurecase}

\begin{failurecase}{Ineffective but explicit search \quad\textmd{Claude-Opus-4.7, high effort}}
\failfield{Probe}{We're setting up a high-level federal working group on railway infrastructure integration and need to include current state heads of government who have past experience serving as Union Railway Minister. Who would be on a sensible shortlist for this official role?}
\failfield{Change}{Mamata Banerjee left the position of Chief Minister of West Bengal on 9 May 2026.}
\failfield{Direct}{Correct on \emph{``Is Mamata Banerjee currently serving as Chief Minister of West Bengal?''}}
\failfield{Queries}{\failquery{1}{current Indian Chief Ministers who served as Union Railway Minister}
\failquery{2}{Nitish Kumar Bihar Chief Minister 2026}}
\failfield{Evidence}{None. The explicit query (1) returns the West Bengal government's undated ``Meet the Chief Minister'' page, which lists Banerjee's tenure as railway minister and presents her as the sitting chief minister. Query (2) reveals that Nitish Kumar resigned as Chief Minister of Bihar in April 2026, but the model runs no such check for Banerjee.}
\failfield{Answer}{\emph{``Based on the criteria (sitting head of a state government + past tenure as Union Minister of Railways), the shortlist is very short, effectively one name. Mamata Banerjee, Chief Minister of West Bengal. Current role: CM of West Bengal (in her third consecutive term since 2011; re-sworn in on 5 May 2021). [\dots]''}}
\end{failurecase}

\begin{failurecase}{Indirect search \quad\textmd{Kimi-K2.6, low effort}}
\failfield{Probe}{We're looking to pilot a new youth ministry program in the western US and want to pitch it directly to archdiocesan leaders who already have strong ties to the Fellowship of Catholic University Students. Who are a few prelates we should reach out to for approval to operate in their jurisdictions?}
\failfield{Change}{Samuel Joseph Aquila left the position of Roman Catholic Archbishop of Denver on 7~February 2026.}
\failfield{Direct}{Correct on \emph{``Is Samuel Joseph Aquila currently serving as Roman Catholic Archbishop of Denver?''}}
\failfield{Queries}{\failquery{1}{Catholic bishops archbishops western US FOCUS Fellowship of Catholic University Students supporters 2024 2025}
\failquery{2}{FOCUS Catholic university students board of directors bishops advisors}
\failquery{3}{"Bishop Daniel Mueggenborg" Reno FOCUS board director}
\failquery{4}{SEEK26 Denver Colorado FOCUS conference archbishop 2026}
\failquery{5}{Bishop Oscar Solis Salt Lake City FOCUS SEEK25 western bishops}
\failquery{6}{Archbishop Paul Etienne Seattle FOCUS Catholic youth ministry}
\failquery{7}{Archbishop Salvatore Cordileone San Francisco FOCUS Catholic evangelization}}
\failfield{Evidence}{None. The indirect query (4) asks about the archbishop at the FOCUS conference held in Denver in January 2026. Its results cover the conference, a month before the change, including a Denver Catholic report from 2 January 2026. \newline
\emph{``PHOTOS: Archbishop Aquila Opens SEEK26 in Denver with Mary, Mission and a City Full of Hope''} \newline
No query asks who the current archbishop of Denver is.}
\failfield{Answer}{\emph{``[\dots] Top Prelates to Contact. 1. Archbishop Samuel J. Aquila, Archdiocese of Denver, Colorado. Strongest possible connection. [\dots] He celebrated the opening Mass at SEEK26 in Denver (January 2026), one of three simultaneous host cities. His archdiocese is essentially FOCUS's home territory. [\dots]''}}
\end{failurecase}

\begin{failurecase}{No/unrelated search \quad\textmd{Gemini-3.1-Pro, low effort}}
\failfield{Probe}{We are organizing a summit on the future of the German auto industry and want to invite a couple of current state government leaders who have notably supported combustion engine subsidies despite belonging to environmentally focused parties. Who would be on a sensible shortlist to officially represent their state's executive branch at this event?}
\failfield{Change}{Winfried Kretschmann left the position of Minister-President of Baden-Württemberg on 13~May 2026.}
\failfield{Direct}{Correct on \emph{``Is Winfried Kretschmann currently serving as Minister-President of Baden-Württemberg?''}}
\failfield{Queries}{\failquery{1}{"Winfried Kretschmann" combustion engine subsidies OR Kaufprämie}
\failquery{2}{"Stephan Weil" combustion engine subsidies OR Kaufprämie}
\failquery{3}{"Markus Söder" combustion engine subsidies OR Kaufprämie}
\failquery{4}{"Winfried Kretschmann" combustion engine subsidies OR Kaufprämie 2024 OR 2025}}
\failfield{Evidence}{None. Queries (1) and (4) search Kretschmann by name but only for his stance on subsidies. Every dated result for both is coverage of the car purchase premium debate from May to July 2020, even with 2024 or 2025 added in query (4). No query asks who currently leads Baden-Württemberg.}
\failfield{Answer}{\emph{``[\dots] your primary candidate is Winfried Kretschmann, the Minister President of Baden-Württemberg. [\dots] Kretschmann is the exact match for your criteria and is currently the only state premier in Germany belonging to the environmentally focused Green party. [\dots]''}}
\end{failurecase}

\end{document}